\newcommand{\ourtitle}{Fairkit, Fairkit, on the Wall, Who's the Fairest of Them All?\\Supporting Data Scientists in Training Fair Models}
\newcommand{\pdftitle}{Fairkit, Fairkit, on the Wall, Who's the Fairest of Them All? Supporting Data Scientists in Training Fair Models}

\documentclass[sigconf,nonacm,screen]{acmart}

\usepackage{balance}       
\usepackage{graphics}      
\usepackage[T1]{fontenc}   
\usepackage{txfonts}
\usepackage{mathptmx}
\usepackage{color}
\usepackage{booktabs}
\usepackage{textcomp}
\usepackage{xcolor}
\usepackage{tcolorbox}

\usepackage{svg}
\graphicspath{{figures/}}

\usepackage{microtype}        
\usepackage{ccicons}          

\usepackage{enumitem}
\newcommand{\subscript}[2]{$#1 _ #2$}

\makeatletter
\def\url@leostyle{%
  \@ifundefined{selectfont}{
    \def\UrlFont{\sf}
  }{
    \def\UrlFont{\small\bf\ttfamily}
  }}
\makeatother
\def\pprw{8.5in}
\def\pprh{11in}

\begin{document}

\title[\pdftitle]{\ourtitle}

\author[Brittany Johnson, Jesse Bartola, Rico Angell, Katherine Keith, Sam Witty, 
Stephen J.~Giguere, and Yuriy Brun]{Brittany Johnson,
Jesse Bartola,
Rico Angell,
Katherine Keith,\\
Sam Witty,
Stephen J.~Giguere,
Yuriy Brun\\
{\small\vspace{1ex} University of Massachusetts Amherst, USA\\ \vspace{-1ex}
\href{mailto:johnsonb@gmu.edu, jrbartola@gmail.com, {rangell, kkeith, switty, sgiguere, brun}@cs.umass.edu}{johnsonb@gmu.edu, jrbartola@gmail.com, \{rangell, kkeith, switty, sgiguere, brun\}@cs.umass.edu}}
}

\begin{abstract}

Modern software relies heavily on data and machine learning, and affects
decisions that shape our world. Unfortunately, recent studies have shown that
because of biases in data, software systems frequently inject bias into their
decisions, from producing better closed caption transcriptions of men's
voices than of women's voices to overcharging people of color for financial
loans. To address bias in machine learning, data scientists need tools that
help them understand the trade-offs between model quality and fairness in
their specific data domains. Toward that end, we present fairkit-learn, a
toolkit for helping data scientists reason about and understand fairness.
Fairkit-learn works with state-of-the-art machine learning tools and uses the
same interfaces to ease adoption. It can evaluate thousands of models
produced by multiple machine learning algorithms, hyperparameters, and data
permutations, and compute and visualize a small Pareto-optimal set of models
that describe the optimal trade-offs between fairness and quality. We
evaluate fairkit-learn via a user study with 54 students, showing that
students using fairkit-learn produce models that provide a better balance
between fairness and quality than students using scikit-learn and IBM AI
Fairness 360 toolkits. With fairkit-learn, users can select models that are
up to 67\% more fair and 10\% more accurate than the models they are likely
to train with scikit-learn.

\end{abstract}

\maketitle

\section{Introduction}

Data-driven software is used increasingly to make automated
decisions that shape our society. Software decides what products
we are led to buy~\cite{Mattioli12};
who gets access to financial instruments~\cite{Olson11} or gets hired~\cite{Raghavan19}; 
what a self-driving car does~\cite{Goodall16},
how medical patients are diagnosed and treated~\cite{Strickland16}, and
when to grant bail~\cite{Angwin16}.

\begin{figure}[t]
\begin{center}
\begin{tabular}{|p{2.4in}|}
\hline
\includegraphics[width=2.4in]{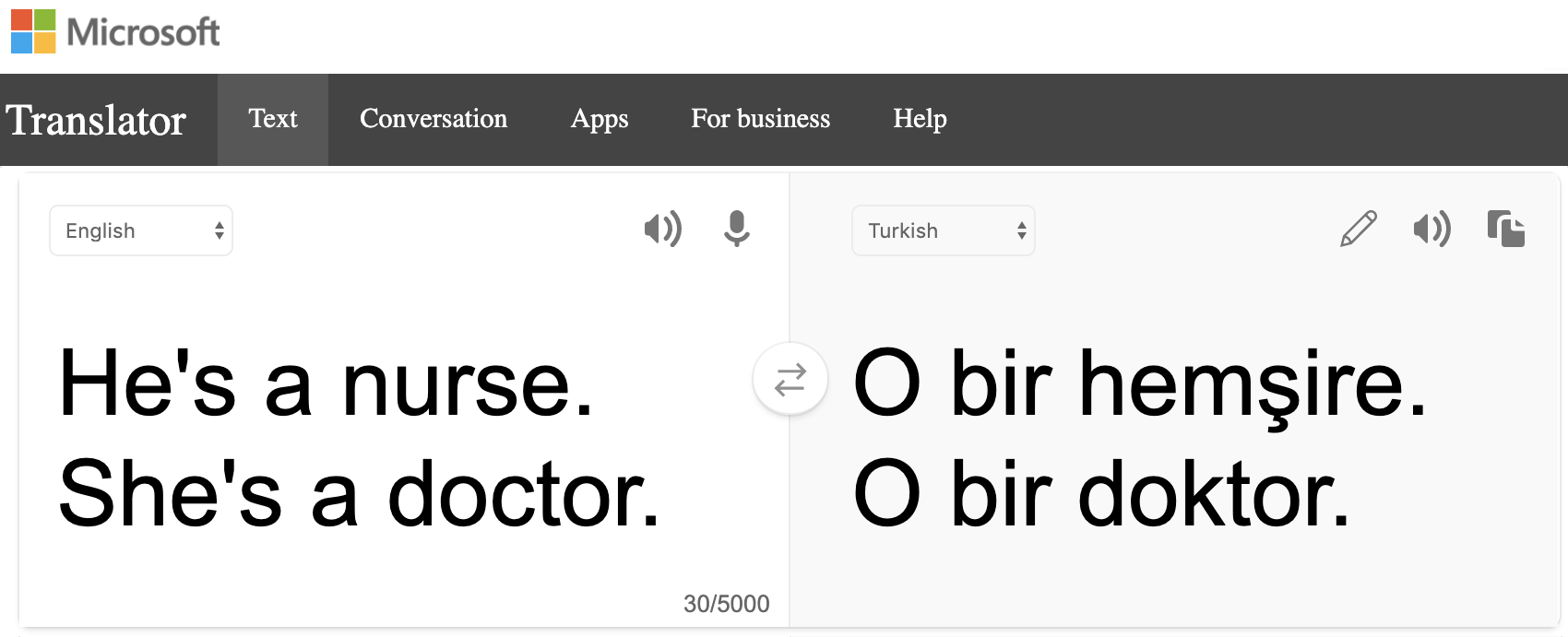} \\
\hline
\includegraphics[width=2.4in]{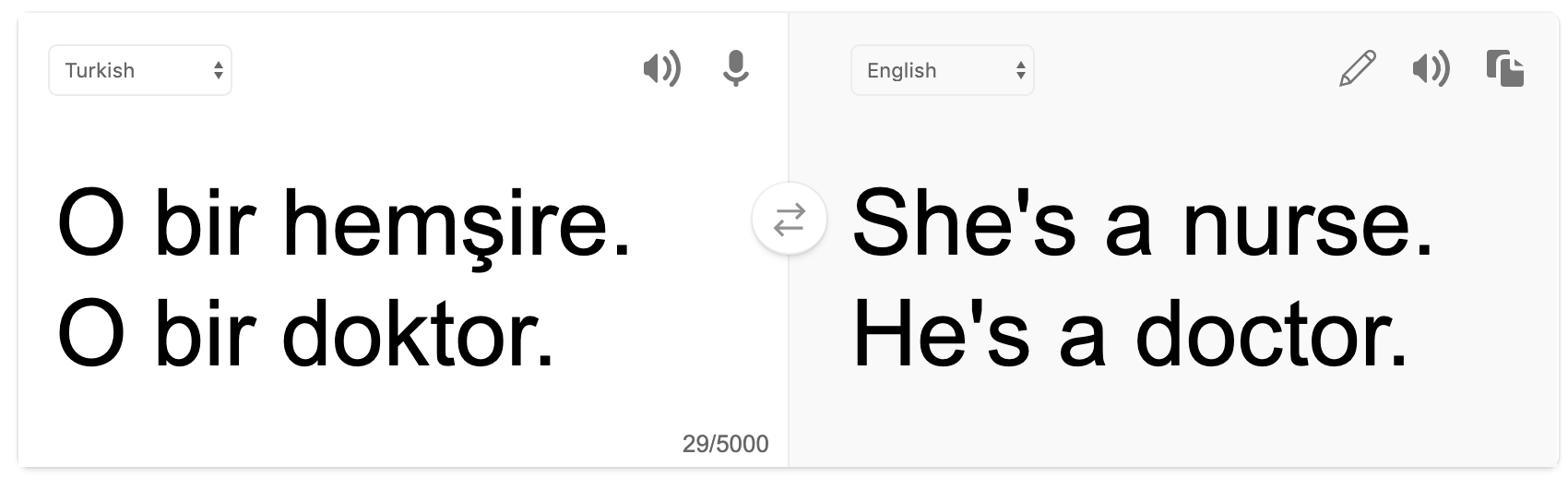} \\
\hline
\end{tabular}
\end{center}

  \caption[Automatically translating text from a language without gendered
  pronouns into English can be influenced by societal biases.]{Automatically
  translating text from a language without gendered pronouns into English can
  be influenced by societal biases~\cite{Caliskan17}.}

\label{fig:translate}
\end{figure}

Unfortunately, recent studies have shown that such software can inherit
biases from data and the environment. For example, translation engines can
inject societal biases: Type ``He is a nurse. She is a doctor.''\ into
\url{https://www.bing.com/translator} and translate it into Turkish. Then
translate the result (``O bir bebek hem{\c{s}}ire. O bir doktor.'')\ into
English and you get ``She is a nurse. He is a doctor.''\ 
(see Figure~\ref{fig:translate}). YouTube makes more mistakes when
automatically generating closed captions for videos with female than male
voices~\cite{Koenecke20, Tatman17}. Racial bias affects the ads search engines display,
e.g., showing ads for (nonexistent) arrest records when searching for African
American names~\cite{Sweeney13}. Amazon's software has failed to offer
same-day delivery to predominantly minority neighborhoods~\cite{Letzter16}, while, 
Staples offered online discounts to customers only in more affluent
neighborhoods~\cite{Mikians12, Haweawar12}. Language processing tools are more
accurate on English written by white people than people of other
races~\cite{Blodgett17}. Facial recognition software recognizes female and
non-white faces less often and less accurately than those of white
men~\cite{Klare12}. And the software US courts use to assess the risk of a
criminal committing another crime exhibits racial bias~\cite{Angwin16}.

One fundamental cause of these biases is that modern software often applies
machine learning to data generated from the real world. First, the real world
is full of biases, often subconscious ones that the people who exhibit them
do not recognize. In fact, humans often do not realize their biased behavior
until they see an automated system reproduce it~\cite{Peng19}. Second,
machine learning is notoriously opaque due to its probabilistic nature,
sensitivity to small design decisions such as hyperparameter tuning, complex
data preprocessing and model architecture, and nontransparent
operation~\cite{Doshi-Velez17, Barocas18, Holstein19}. As a result, models learned from
data can often encode discriminatory behavior from the data's bias, but that
behavior is both hard to identify and eliminate~\cite{Galhotra17fse}.

Recently, public's demand for transparency in data and learned model use has
increased~\cite{albrecht2016gdpr}, and governments have initiated efforts to
increase regulation of decisions made by software systems to reduce bias and
improve transparency~\cite{Soper16NYChicago, President16, deBlasio18}. 
As a result, it is increasingly important to
provide support tools for those who apply machine
learning to data, study data, and build software systems that use data to
make decisions. These tools must support detecting and understanding biases
in data and learned models, and the inherent trade-offs between mitigating
bias and maximizing decision quality. 
Industry experts have called for tools that help data scientists understand
bias in data and curate datasets, and to audit and debug fairness
issues~\cite{Holstein19}, all of which our paper aims to address.

The challenges in helping data scientists reason about fairness include:
(1)~Fairness (as well as quality) mean different things in different data
domains, and no single definition of fairness is universally appropriate,
with definitions often being mutually exclusive on datasets. (2)~The
trade-offs between fairness and quality are typically a function of the data
and not of the tools applied to train models, and algorithms that produce
fair models on some datasets may produce biased ones on others. (3)~The space
of possible models machine learning can produce is astronomically large due
to the combinatorial explosion caused by a large number of learning
algorithms, hyperparameters, and data permutations that affect the models.
(4)~Learning algorithms that attempt to account for fairness typically do not
provide guarantees on the behavior of the models they produce~\cite{Zafar15a,
Zafar17}, and can sometimes inject more bias than fairness-unaware
algorithms~\cite{Galhotra17fse}; using fairness-aware algorithms to reduce
one kind of bias can significantly increase other
biases~\cite{Galhotra17fse}; and learning algorithms that do provide
guarantees about their models' fairness can, under some conditions, break
those guarantees~\cite{Agarwal18} or fail to produce a model altogether, even
if fair ones exist~\cite{Thomas19science, Metevier19neurips}.

\begin{figure*}[t]
\centering
  \def\svgwidth{\textwidth}
  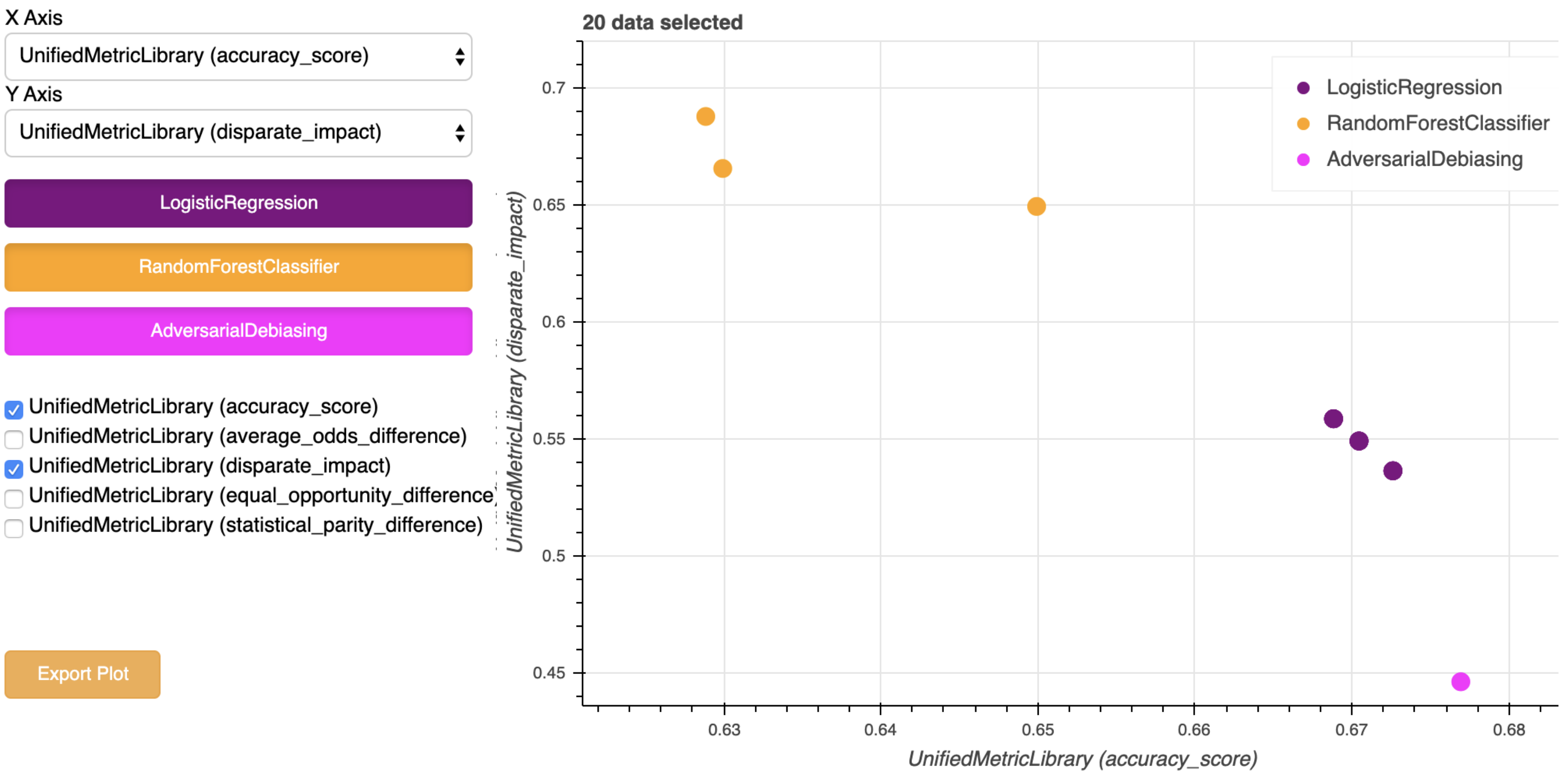
  \caption[Fairkit-learn trains and evaluates a large-number of machine
  learning models using multiple learning algorithms]{Fairkit-learn trains
  and evaluates a large-number of machine learning models using multiple
  learning algorithms (here, logistic regression, random forrest, and
  adversarial debiasing) and an array of hyperparameters, finding the
  Pareto-optimal set of models that represent the best combination of quality
  metrics (here, accuracy, shown on the y-axis) and fairness metrics (here,
  disparate impact, shown on the x-axis). Faitkit-learn's visualization helps
  data scientists understand the domain of their data (here, the COMPAS
  recidivism dataset~\cite{Angwin16}), explaining relationships and
  trade-offs between quality and fairness metrics, and showing which
  algorithms achieve better combinations of multiple metrics.}
  \label{fig:results_plot}
\end{figure*}

While some modern tools can measure various dimensions of fairness of a given
model~\cite{aif360, Bellamy16, Galhotra17fse, tramer2017fairtest, Adebayo16},
and some machine learning algorithms can train models while enforcing
fairness constraints~\cite{Thomas19science, Agarwal18, Metevier19neurips, aif360, Zafar15a,
Zafar17}, none of these tools provide support for understanding the
trade-offs between fairness and quality of the models and for comparing and
contrasting models along the combinations of fairness and quality measures
they produce. For example, scikit-learn, the state-of-the-art go-to toolkit
used ubiquitously by data scientists in industry~\cite{scikitlearn}, provides
tools for training many types of machine learning models, and evaluating them
for quality, such as precision and recall, but not fairness metrics. IBM's open source toolkit, 
AI Fairness 360, adds support for computing
fairness metrics on learned models and learning algorithms that account for
some definition for fairness~\cite{aif360, Bellamy16}. Fairness-aware
learning algorithms, such as fairlearn~\cite{Agarwal18} or
RobinHood~\cite{Metevier19neurips} and others
designed using the Seldonian Framework~\cite{Thomas19science}, can enforce
fairness constraints, but
without helping data scientists understand how that enforcement affects model
quality. The bottom line is, these tools still fail to provide support for
understanding the trade-offs between fairness and quality, e.g., to help data
scientists answer questions such as ``Does finding a more fair model
necessarily imply the model's quality will decrease, and by how much?''

Toward this end, we have developed fairkit-learn, a tool that builds on
scikit-learn and IBM AI Fairness 360, to help data scientists better
understand the model fairness landscape. Fairkit-learn uses the same
interface as scikit-learn, easing adoption, and works with all of
scikit-learn's and AI Fairness 360's algorithms, metrics, and datasets, and
provides interfaces for easily including more definitions, metrics, and
datasets. Fairkit-learn uses visualization to help data scientists understand
the fairness properties of specific models, which learning algorithms learn
models that better satisfy competing requirements of fairness and quality in
a particular domain, and demonstrate opportunities for selecting models that
improve fairness or quality at the lowest expense of the other. For example,
fairkit-learn can perform a grid search through tens of thousands of possible
models learned using different machine learning algorithms with different
combinations of hyperparameters, and select the Pareto-optimal set of models
with respect to multiple data-scientist-selected fairness and quality
definitions.

In addition to combining multiple fairness and quality considerations,
allowing data scientists to optimize them simultaneously, fairtkit-learn
provides four other benefits over prior work. First, fairkit-learn simplifies
the process of exploring the space of possible models by automatically
performing grid searches over multiple learning algorithms, model
hyperparameters, and data permutations, lifting the burden of
implementing such a search off the user. Second, unlike existing tools that audit learned 
models for fairness~\cite{Galhotra17fse, tramer2017fairtest,
Adebayo16}, faitkit-learn helps data scientists understand the
fairness-quality trade-off landscape \emph{during} model development,
allowing them to make design decisions about these trade-offs when those
decisions can still affect overall system performance. Third, fairkit-learn
works with dozens of different definitions of fairness and quality, allowing
the data scientists to evaluate the applicability of different definitions to
their data domains and select those that make most sense in their particular
situations. Fourth, fairkit-learn uses a visualization-based approach to
clearly communicate trade-offs to the data scientists, helping them make
informed decisions.

Figure~\ref{fig:results_plot} shows a sample fairkit-learn visualization.
Here, a data scientist is comparing models learned by three learning
algorithms\,---\,logistic regression, random forest classifier, and
adversarial debiasing\,---\,on the COMPAS recidivism dataset. 
Fairkit-learn trains approximately 80 different 
models using these three algorithms by varying their hyperparameters in a
grid-search, and computes the much smaller (here, seven) subset of the models
that make up the Pareto-optimal set. Fairkit-learn visualizes the seven
models with respect to two metrics selected by the data scientist: disparate
impact, a fairness metric, visualized on the x-axis, and model accuracy, a
quality metric, visualized on the y-axis. The visualization elides
multiple sub-optimal models to show only those for which improving
fairness decreases accuracy, and vice versa (the Pareo-optimal model set).
This visualization makes it easy to see that (1)~in
this data domain, model fairness and model accuracy are opposing forces (in
other domains, they can be complementary), (2)~a small reduction in
quality (63\% versus 68\%) can produce a large increase in fairness (69\%
versus 45\%), and (3)~random forest classifier models (orange) tend to
produce more-fair models at a slight cost in accuracy, adversarial debiasing
(magenta), a machine learning algorithm intended to be fairness-aware, produces
less fair but slightly more accurate models, and logistic regression (purple)
models perform slightly better than adversarial debiasing.

We evaluate fairkit-learn in a controlled user study with 54 students
studying data science and software engineering.\footnote{This human-subject
study was approved by the Institutional Review Board.} Our within-subject
study asked subjects to use scikit-learn, IBM AI Fairness 360, and
fairkit-learn to explore the machine-learning-model landscape on three
datasets, aiming to produce models that satisfy a combination of fairness and
quality metrics. We found that subjects who used fairkit-learn produced more
fair models than when using scikit-learn and that while IBM AI Fairness 360
may be better for data scientists only interested in improving fairness,
fairkit-learn supports finding models that are both fair and high quality
(more so than AI Fairness 360). With fairkit-learn, users can select models that are
up to 67\% more fair and 10\% more accurate than the models they are likely
to train with scikit-learn.

The main contributions of this work are:
\begin{itemize}

\item  Fairkit-learn, a novel open-source tool that uses familiar interfaces and visualization for exploring, evaluating, and visualizing
the performance and fairness trade-offs in machine learning models.

\item A user study, evaluating fairkit-learn against scikit-learn and IBM
AI Fairness 360, showing that subjects using fairkit-learn train models that better 
balance fairness and accuracy. Our study also provides insights into how data scientists reason about
fairness when using traditional machine learning tools, e.g., scikit-learn,
to train and evaluate models.

\end{itemize}

\section{The fairkit-learn toolkit}

Fairkit-learn is an open-source, publicly available Python toolkit designed
to help data scientists evaluate and explore machine learning models with
respect to quality and fairness metrics simultaneously.\footnote{Link to
fairkit-learn is redacted for double-blind review.}

Fairkit-learn builds on top of scikit-learn, the state-of-the-art tool suite
for data mining and data analysis, and AI Fairness 360, the state-of-the-art
Python toolkit for examining, reporting, and mitigating machine learning bias
in individual models~\cite{aif360, scikitlearn}. Fairkit-learn supports all
metrics and learning algorithms available in scikit-learn and AI Fairness
360, and all of the bias mitigating pre- and post-processing algorithms
available in AI Fairness 360, and provides extension points to add more
metrics and algorithms.

This section describes the complexity of model fairness and the space of
fairness definitions fairkit-learn handles, fairkit-learn's search
capabilities for helping data scientists explore and understand the space of 
possible models and data permutations, and fairkit-learn's analysis
of Pareto-optimal sets of models and visualization capabilities for
illustrating trade-offs between model fairness and quality.

\subsection{Integrated machine learning tools}

We selected two existing machine learning toolkits as the foundation for fairkit-learn: scikit-learn~\cite{scikitlearn} and IBM's AI Fairness 360~\cite{aif360}. We discuss each of these tools separately below.

\subsubsection{Scikit-learn}

Scikit-learn is a commonly used and integrated machine learning toolkit,
therefore we wanted to ensure that fairkit-learn works with its models and
functionality. While scikit-learn provides a number of algorithms and metrics
for training and evaluating machine learning models, it does not support
training or evaluating models for fairness. It also does not have built-in
support for exploring the space of machine learning model configurations; if
a data scientist wants to find an optimal model for a given metric, she must
implement the code to do so herself. Scikit-learn also only supports
evaluating machine learning models by one metric at a time\,---\,any
trade-off analysis has to be written by the user.

\subsubsection{AI Fairness 360}

IBM AI Fairness 360 provides an exhaustive set of datasets, models, algorithms, and metrics that pertain to machine learning model fairness, so we used this toolkit as the foundation for fairkit-learn's fairness components. Along with this large set of functionalities, the website provides detailed documentation and examples for using the various components of the toolkit.~\footnote{https://aif360.mybluemix.net} And like fairkit-learn, AI Fairness 360 is built using scikit-learn. However, AI Fairness 360, like scikit-learn, does not provide built-in support for exploring the space of models and configurations nor does it provide support for evaluating trade-offs between multiple metrics. Any trade-off evaluations, along with model configuration exploration, would have to be implemented by the user.

\subsection{Fairness metrics}

Fairness is a broad notion that can be partially represented by many
formal definitions~\cite{Narayanan18}. Unfortunately, users, data
scientists, and regulators rarely agree on a single
definition~\cite{grgic2018human}, though they often agree that fairness, in
some form, is important~\cite{woodruff2018qualitative}. In fact, while each
definition of model fairness is appropriate in some context,
many are impossible to satisfy simultaneously~\cite{Kleinberg17, Friedler16}.
To effectively support data scientists across many domains,
\underline{fairkit-learn} \underline{supports many fairness definitions, including all 
supported by IBM} \newline \underline{AI Fairness 360~\cite{aif360, Bellamy16}}, and provides
extension points to add more.

Here, we describe several representative definitions of fairness
fairkit-learn currently handles to give the reader as sense of their diversity. More
complete lists exists, e.g.,~\cite{Narayanan18}, and active research in
the area of fairness definitions is continually expanding that list at this
time.

\begin{itemize}

\item \textbf{Disparate treatment} is a concept originally of legal origins.
The computer science formalization of this definition says that for a model
to satisfy disparate treatment with respect to a set of attributes, it must
have been learned without access to those attributes~\cite{Zafar17}. Note,
however, that this definition often fails to ensure meaningful fairness in
practice, because data attributes are often correlated, e.g., age correlates
with savings, race correlates with name, and, in the United States, race
correlates with zip code, models trained without access to a set of
attributes can still effectively act unfairly with respect to those
attributes~\cite{Sweeney13, Ingold16}.

\item \textbf{Disparate impact} captures the notion that a model may have
adverse effects on protected groups~\cite{GriggsVDukePowerCo,
Chouldechova17, Zafar17}. To satisfy the disparate treatment
definition, a model must treat similarly the same fraction of individuals of
each group. For example, if an employer hires $\frac{1}{2}$ of its male
applicants, then that employer must hire at least $\frac{1}{2}$ of its female
applicants~\cite{GriggsVDukePowerCo}. If the fractions are different, the
ratio between the fractions is a measure of bias.

\item \textbf{Demographic parity}, also called \textbf{statistical parity}
and \textbf{group fairness}, is closely related to disparate impact, and
requires that the model's predictions are statistically independent of the
attribute with respect to which the model is fair~\cite{Dwork12,
Calders10}. The measure of bias is, unlike for disparate impact, the
difference between the fractions.

\item \textbf{Delayed impact} is concerned with the fact that making
seemingly fair decisions can, in the long term, produce unfair
consequences~\cite{Liu18}. For example, to make up for a disparity in
recidivism predictions by race, a model may, at random, decrease its
predictions for one race. While on its face, this may improve the situation
for members of that race, if this results in more visibility for repeat
offenders of that race, the public's perception may have a more negative
effect toward that race, producing delayed negative impact. Measuring delayed
impact requires temporal indicator data, of, for example, long-term
improvement, stagnation, and decline in variables of interest~\cite{Liu18}.

\item \textbf{Predictive equality} requires that false positive rates are
equal among groups~\cite{Chouldechova17, corbett2017algorithmic}.

\item \textbf{Equal opportunity} requires that false negative rates are equal
among groups~\cite{Hardt16, Chouldechova17}.

\item \textbf{Equalized odds}, a combination of predictive equality and equal
opportunity, requires that both false positive and false negative rates are
equal among groups~\cite{Hardt16}. Consequently, the equalized odds
criterion can be viewed as the conjunction of the predictive equality and
equal opportunity criteria.

\item \textbf{Treatment equality} requires that the ratio of the
false-positive rate to the false-negative rate is the same for each
group~\cite{Berk17}.

\item \textbf{Causal fairness}, also called \textbf{counterfactual fairness},
is based on the counterfactual causal relationship between variables. To be
causally fair, a classifier must predict the same label for all feature
vectors that are the same except for those attributes. In other words, if two
individuals differ only in protected attributes, and are otherwise identical,
this definition requires classifiers to predict the same outcome for both
individuals~\cite{Galhotra17fse, Kusner17}. For example, a recidivism model
is causally fair with respect to race only if it predicts identical labels
for all pairs of individuals identical in every way except race.

\item \textbf{Metric fairness} requires that, given a distance metric to
compare two feature vectors, the model should predict similar labels for
similar feature vectors, on average~\cite{Dwork12}. \textbf{Approximate
metric fairness} extends this definition by incorporating a tolerance
parameter to obtain generalization bounds~\cite{Rothblum18}.

\item \textbf{Representation disparity} limits the error for all
subgroups~\cite{Hashimoto18}. The amount of representation
disparity is the maximum loss for any particular group.

\item \textbf{Conditional use accuracy equality} requires that precision (the
probability that the model is correct when it predicts a label) is the same
for all groups~\cite{Berk17}.

\item \textbf{Overall accuracy equality} requires that the accuracy of the
classifier (fraction of the feature vectors that the model correctly
classifies) is equal for each group~\cite{Berk17}.

\end{itemize}

\subsection{Model search}

\begin{figure}
\centering
  \includegraphics[width=3.5in]{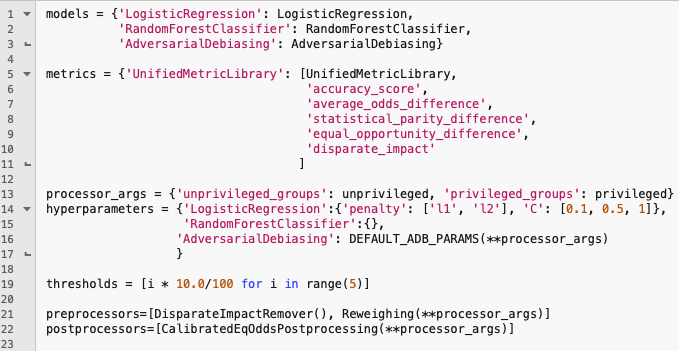}
  \caption{Example parameters for model search in fairkit-learn}
  \label{fig:search_params}
\end{figure}

Unlike existing tools, which require data scientists to write their own code to evaluate more than one model configuration, fairkit-learn provides functionality that allows data scientists to search over any number of model configurations (given enough memory and power) for pareto optimal solutions (that best balance quality metrics of concern and fairness). Figure~\ref{fig:search_params} shows code that initializes each parameter required for the model search: \emph{models}, \emph{metrics},
\emph{hyperparameters}, \emph{thresholds}, and \emph{pre-/post-processing algorithms}.

\subsubsection{Models}
To run the grid search, you need to specify at least one model to include (\texttt{models} in Figure~\ref{fig:search_params}). You can specify as many models as available computational resources will allow. Fairkit-learn is currently compatible with scikit-learn and AI Fairness 360, but can be extended to work with others via the model wrapper class provided.

\subsubsection{Metrics}
Also required for the grid search are metrics to evaluate each model configuration (\texttt{metrics} in Figure~\ref{fig:search_params}). Fairkit-learn is currently compatible with metrics from scikit-learn and AI Fairness 360, but uses a wrapper metric class that can be extended with other metrics.

\subsubsection{Hyperparameters}
Fairkit-learn can run using default hyperparameters, or users can provide different values for each hyperparameter to evaluate in the grid search. The example in Figure~\ref{fig:search_params} runs the \texttt{AdversarialDebiasing} and \texttt{RandomForestClassifier} models with default parameters and provides options for two of the \texttt{LogisticRegression} model hyperparameters.

\subsubsection{Thresholds}
The threshold parameter denotes the probabilistic threshold required to be considered a positive classification (in a binary classification). For example, if the threshold is 0.7, then any prediction with $\geq 0.7$ probability will be considered favorable.

\subsubsection{Pre- and post-processing algorithms}
Finally, users have the option of specifying any data pre-processing or model post-processing algorithms to include in the search (\texttt{preprocessors} and \texttt{postprocessors} in Figure~\ref{fig:search_params}). Fairkit-learn currently works with pre- and post-processing algorithms provided by AI Fairness 360.

Once the search is done, results are written to a .csv file. The .csv file is used to render a visualization of the results of the grid search.

\subsection{Search result visualization}

To help data scientists process the results of the grid search, fairkit-learn provides functionality that allows users to visualize the results. The visualization shown in Figure~\ref{fig:results_plot} is showing some results from the search shown in Figure~\ref{fig:search_params}. More specifically, the visualization is showing the \emph{Pareto frontier} of the   \texttt{LogisticRegression}, \texttt{RandomForestClassifier}, and  \texttt{AdversarialDebiasing} models with respect to accuracy and (\texttt{accuracy\_score}) and fairness (\texttt{disparate\_impact}). 

When using the fairkit-learn visualization, one can view the Pareto frontier of any two metrics by selecting those metrics in the checklist and for the X and Y axes (as shown in Figure~\ref{fig:results_plot}). To access all search results (including not Pareto optimal), select all metrics and choose the X and Y axis you want to view. Engineers can also toggle which models to show in the plot by clicking the model button (e.g., the \textcolor{magenta}{magenta} \texttt{AdversarialDebiasing} button) hover over the data points in a given plot to get more information on the model configuration at that point (e.g., hyperparameter values). The visualization can also be exported for later viewing and comparison, along with a JSON file that describes the exported plot.

\section{Evaluation}

To evaluate fairkit-learn and explore how data scientists train fair models, we conducted a user study to validate the following hypotheses:

\begin{enumerate} [label=\subscript{H}{{\arabic*}}]
	\item Compared to out-of-the-box scikit-learn models, fairkit-learn supports training fairer models.
	\item When asked to find the most fair model, individuals that use fairkit-learn are able to train models that are more fair.
	\item When asked to find a model that best balances fairness and accuracy, individuals that use fairkit-learn are able to train models that are more fair and comparably accurate.
\end{enumerate}

We also collect data to answer the question \emph{how do data scientists reason about model fairness when not using fairness tooling?}

\subsection{Datasets}
\label{subsec:datasets}

We used three real-world datasets to evaluate fairkit-learn. Each dataset has
its own definition of which groups are privileged and can be used for binary
classification tasks.

\subsubsection{ProPublica COMPAS dataset}

The COMPAS dataset is publicly available and contains recidivism data for
defendants in Browards County between 2013 and 2014~\cite{compasdata}. For
each individual in the dataset, the dataset includes their criminal history
both before and after arrest, and the risk assessment score, as calculated by
the COMPAS system~\cite{Angwin16}. In 2016, ProPublica found significant
differences between predictions the COMPAS system made based on race, finding
that the system more often predicted African American defendants would commit
a crime again, when, in reality, they did not, while predicting that white
defendants would not a commit a crime again, when, in reality, they did. Data
scientists can use this dataset to train models to predict recidivism and to
ensure fairness. For our analyses, we treated Caucasian females as the the
privileged group, treating race and sex as protected attributes.

\subsubsection{German credit dataset}

The German credit dataset is publicly available and contains financial data
of 1,000, some of whom are classified as potential credit
risks~\cite{statloggermancreditdata}. The dataset consists of attributes
ranging from credit history to personal status and sex. Data scientists
can use the German credit dataset when, for example, training models for use
in banking or loan approval software. For our analyses, we treated the men 25
years of age or older as the privileged group, treating age and sex as
protected attributes.

\subsubsection{Adult census income dataset}

The Adult census income dataset is publicly available and contains Census
data, such as race, occupation, and salary, for 48,842 individuals from
1994~\cite{adultincomedata}. Data scientists can use this dataset to train
models that make income predictions (e.g., whether a person make more than
US\$50K per year). For our analyses, we treated Caucasian men as the
privileged group, treating sex and race as the protected attributes.

\subsection{User study design}

To validate our hypotheses and answer our research question, we designed a user study to explore the effects of various tooling on the machine learning models data scientists train. The state-of-the-art in training and evaluating machine learning models, and at the core of both fairkit-learn and AI Fairness 360, is scikit-learn.  Therefore, we designed our experiment with two control groups: one that only uses functionality provided by scikit-learn and the other only using AI Fairness 360.

To make our study design more realistic, we created a Jupyter notebook\footnote{\url{https://jupyter.org/}} for each experimental group.
We presented the notebooks to participants as a homework assignment with three tasks.

Each notebook provided information on the tasks, relevant details, and links to external documentation.
For each task, we provided participants with a real-world dataset and a tutorial on how to use one of the tools. Following each tutorial, we asked them to complete the following subtasks:

\begin{enumerate}

  \item Find a machine learning model you believe will be the most accurate.

  \item Find a machine learning model you believe will be the most fair.

  \item Find a machine learning model you believe will best balance both accuracy and fairness.

\end{enumerate}

For our evaluation, we selected a subset of the metrics available for use in fairkit-learn. However, as previously mentioned, users of fairkit-learn can incorporate and use any fairness metric of their choice. 
To complete the tasks, we gave each participant all the necessary study materials and instructions for participation.
The first task notebook provided participants with background information on what they would be doing, the tools they will be using, and where to submit their responses.

Our design consisted of 6 experimental groups. To reduce the effects of learning bias on the validity of our findings, each experimental group used the tools in a different order. We used the same three datasets 
from AI Fairness 360 for all 6 experimental groups. This allowed us to
increase confidence that our findings generalize.
 
After the exercise was complete, we collected participant notebooks and other relevant data. Next, we outline what data we collected and how.

\subsection{Data collection}

We collected data from participants as they completed the exercise in an online response form. 
The form consisted of 5~pages. The first page asked participants demographic questions, including questions about their experience with Python and various Python machine learning tools. 
The next three pages corresponded to each task where we asked participants variations of the following questions, depending on the subtask:

\begin{enumerate}
	\item \emph{Describe the best model and report its metric(s) scores.}
	\item \emph{Why did you select this model?}
\end{enumerate}

We also collected notebook changes and snapshots using nbcomet,\footnote{\url{https://github.com/activityhistory/nbcomet}} an open source tool for tracking jupyter notebook changes. We used this data to triangulate with form responses when possible and necessary.

\subsection{Data analysis}

After data collection, we had to clean, prepare, and analyze the data. The first step in our data analysis was to extract and clean responses from the response form.
We first had to extract, organize, and make each participants' responses anonymous. 
We organized participant responses by task, and then by tool within each task, since that is how we planned to analyze the data.

Research has shown that one of the challenges data scientists have is dealing with the machine learning aspects of working with data, such as feature engineering and hyperparameter tuning~\cite{sanders2017informing,kim2017data}. Therefore, under the assumption that not many data scientists do much if any parameter tuning, to evaluate \subscript{H}{1} we compared each of the scikit-learn models from our study with default parameter settings to the models participants selected.  For each default model, we calculated accuracy and fairness scores for each fairness metric used in the study. We then compared the averages of each for the default models to the averages for participant models selected when using fairkit-learn for fairness related sub-tasks.

To evaluate \subscript{H}{2} and \subscript{H}{3}, we calculated fairness scores and accuracy for each of participants' model selections in the ``find the most fair model'' and ``find the model that best balances both'' sub-tasks. We averaged scores for each metric and measured the difference between those averages using a two-sample t-test ($\alpha=0.05$). 

To answer \emph{RQ1}, we analyzed responses from the tasks where participants used scikit-learn to find the most fair model. We extracted model selections and the qualitative and quantitative rationales for fair model selections in the response form. We then categorized the methods used by participants into the following categories: (1)~\emph{did not try to evaluate for fairness}, (2)~\emph{evaluated with a metric}, (3)~\emph{evaluated with something other than a metric}, and finally (4)~\emph{implemented one or more fairness metrics for evaluation}. We kept track of metrics, and other information, used when participant did try to evaluate fairness.

\subsection{Participants}

We recruited 54 participants from an advanced software engineering course: 30 undergraduates and 24 graduate students. One participant reported having industry experience as a data scientist.

Twenty-six participants had experience with using scikit-learn prior to participating. One participant had prior experience using AI Fairness 360 and no participants had prior experience with fairkit-learn. Fifty-one participants had experience with other Python data science and machine learning tools, such as numpy, scipy, and tensorflow. Fourty participants had prior experience using jupyter notebooks.
On average, participants had approximately 2 years of Python programming experience; this excludes two participants who did not report any years of experience with Python despite reporting having experience with various Python tools.

\section{Results}

We used data collected from our user study to validate and explore how data scientists train models and evaluate them for fairness and accuracy.
In comparing fairkit-learn models to scikit-learn default models, we found that even with concerns split between fairness and accuracy, participants selected fairer, more accurate models when using fairkit-learn (Figure~\ref{fig:default_comparison}).
When fairness is the only concern, our study found that AI Fairness 360 can generally find more fair models than fairkit-learn and scikit-learn (Figure~\ref{fig:fairness_model_scores}).
When trying to balance fairness and accuracy, fairkit-learn is capable of finding models that are high performing and generally more fair than models found by AI Fairness 360 and scikit-learn (Figure~\ref{fig:balance_model_scores}. 

When evaluating fairness without using tools designed to do so, our study found that data scientists have different ways of reasoning about model fairness ranging from ``educated guesses'' to implementing their own fairness metrics. Most often, participants use accuracy as some proxy for fairness. However our data suggests data scientists may have different ideas of the relationship between accuracy and fairness.

\begin{figure*}[ht]
\centering
  \begin{tabular}{lccccc}
	\toprule
  	\textbf{Tool} 							& \textbf{Average Odds} 	& \textbf{Statistical Parity} 	& \textbf{Equal Opportunity}  &\textbf{Disparate Impact}  &  \textbf{Accuracy} \\
  	\midrule
  	scikit-learn (default)     			 & 0.173                    			& 0.221                   				& 0.150    								& 0.555 								& 0.741 \\
	fairkit-learn (fairness)			 & 0.116               					& 0.225                  				& 0.093    								& 0.725 								& 0.788\\
  	fairkit-learn (fairness+accuracy)  & 0.086  						& 0.229                		  			& 0.070    								& 0.829 								& 0.815\\
	\bottomrule
  \end{tabular}
  \caption{Average fairness scores across tasks of default scikit-learn models and fairkit-learn models selected by participants for fairness related subtasks.}
  \label{fig:default_comparison}
\end{figure*}

\begin{figure*}[ht]
	\centering
	\begin{tabular}{lccccc}
		\toprule
		\textbf{Tool} 				& \textbf{Average Odds} 	& \textbf{Statistical Parity} 	& \textbf{Equal Opportunity}  &\textbf{Disparate Impact}  &  \textbf{Accuracy} \\
		\midrule
		scikit-learn      			 & 0.163                    			& 0.213                   				& 0.154    								& 0.570 								& 0.734 \\
		AI Fairness 360			 & 0.079               					& 0.200                  				& 0.061    								& 0.814 								& 0.816 \\
		fairkit-learn  				& 0.116  								& 0.225                		  			& 0.093  								& 0.725 								& 0.788 \\
		\bottomrule
	\end{tabular}
	\caption{Average fairness scores across tasks of models selected by participants for fairness subtasks.}
	\label{fig:fairness_model_scores}
\end{figure*}

\begin{figure*}[ht]
	\centering
	\begin{tabular}{lccccc}
		\toprule
		\textbf{Tool} 				& \textbf{Average Odds} 	& \textbf{Statistical Parity} 	& \textbf{Equal Opportunity}  &\textbf{Disparate Impact}  &  \textbf{Accuracy} \\
		\midrule
		scikit-learn      			 & 0.190                    			& 0.242                   				& 0.171    								& 0.533 								& 0.747 \\
		AI Fairness 360			 & 0.097               					& 0.258                  				& 0.076    								& 0.820 								& 0.822 \\
		fairkit-learn  				& 0.086  								& 0.229                		  			& 0.070    								& 0.829 								& 0.815 \\
		\bottomrule
	\end{tabular}
	\caption{Average fairness scores across tasks of models selected by participants for balancing fairness and accuracy subtasks.}
	\label{fig:balance_model_scores}
\end{figure*}

\subsection{\subscript{\textbf{H}}{\textbf{1}}: Out-of-the-box  model fairness}
One of the most commonly used toolkits for machine learning is scikit-learn, and it is not clear how often data scientists modify default parameters. One of the main advantages to using fairkit-learn is that it support searching different hyperparameter configurations with less effort and expertise. 
As shown in Figure~\ref{fig:default_comparison}, participant models selected for fairness related subtasks while using fairkit-learn outperformed all scikit-learn default models with respect to all metrics considered. We see even more improvement when participants used fairkit-learn to find models that best balance between fairness and accuracy, with the differences as high as 0.30 between fairkit-learn and scikit-learn model fairness scores. These fairness improvements came with no sacrifice to accuracy; in fact, we see an increase of 0.04--0.07 in accuracy score.

\begin{tcolorbox}
	\noindent
	Our findings suggest that when using fairkit-learn, data scientists can
	find models that are more fair and accurate than out-of-the-box, default scikit-learn models.
\end{tcolorbox}

\subsection{\subscript{\textbf{H}}{\textbf{2}}: More fair models with fairkit-learn}

One of the goals of fairkit-learn is to help data scientists find the
fairest models possible. When asked to find a model that will be the most
fair, overall the models selected by participants using fairkit-learn were
more fair than models selected using scikit-learn.
While comparable, the models selected when using AI Fairness 360 were generally more fair than models selected when using fairkit-learn.
Figure~\ref{fig:fairness_model_scores} shows the average fairness scores by metric across tasks by tool.
While one of the goals of fairkit-learn is to support finding fair models, the primary goal is to support balancing fairness with other quality concerns. Given the primary goal of AI Fairness 360 is to support training fair models, these findings are not surprising.

\begin{tcolorbox}%
  \noindent
  Our findings suggest that when using fairkit-learn, data scientists can
  find models that are comparable in fairness to those found using AI Fairness 360 and more fair than models found using scikit-learn.
\end{tcolorbox}

\subsection{\subscript{\textbf{H}}{\textbf{3}}: Fair and accurate models with fairkit-learn}

While fairkit-learn wants to help data scientists improve the fairness of their models, the primary goal of our toolkit is to  support data scientists when attempting to balance fairness with other quality metrics, like accuracy.
Figure~\ref{fig:balance_model_scores} shows the average fairness by metric and accuracy of models selected as the most balanced across tasks by tool.
When asked to find a model that will best balance fairness and accuracy, participants selected models that were more fair with almost no loss with respect to accuracy across all tasks when using fairkit-learn. When analyzing each task individually, we found greater improvement with respect to fairness over scikit-learn when using fairkit-learn than when using scikit-learn. For Task 3, fairkit-learn models were more fair as well as more accurate than AI Fairness 360 or scikit-learn models.

\begin{tcolorbox}%
Our findings suggest that data scientists find more fair models of
high quality when using fairkit-learn than when using AI
Fairness 360 or scikit-learn. When using fairkit-learn, the balanced models
are comparably accurate, and sometimes more accurate, compared to using AI Fairness 360 or scikit-learn.
\end{tcolorbox}

\subsection{Evaluating fairness without fairness tools}

We expect that data scientists would be able to at least reason about fairness when using tools like fairkit-learn and AI Fairness 360, given they provide functionality for doing so. It is less obvious how data scientists handle fairness when using tools like scikit-learn that do not provide functionality for training or evaluating fair models.

For four participants, lack of immediate or easy access to fairness tooling rendered them either unable to find a model they felt would be fair or unable to reason about why a given model should be considered fair.
When asked to select a model they felt would be most fair while using scikit-learn, all participants selected a model. However, the ability to explain why they selected that model over others varied. For Task~1, two participants could not figure out how to reason about the fairness of the model they selected. P47, rather than using some metric or resource to reason about the fairness of their model, put ``not applicable.''

For those who did try to reason about the fairness of their models, participants used various (and sometimes contradictory) ways of evaluating the fairness of a given model. Only four participants used what would be considered fairness metrics to evaluate their models. Two participants used another fairness tool, FairML~\cite{adebayo2016fairml}, to evaluate model fairness. Seven participants created their own metric to evaluate model fairness.

The majority of participants that used a metric (21 out of 25) used model accuracy, or some related metric, to evaluate fairness.
However, participants were split on whether higher accuracy was an indicator of being more fair or if lower accuracy was a better indicator.
Eight out of 21 participants that used accuracy reported selecting a given model because it has the highest accuracy. But three out of the 21 participants felt that \emph{lower} accuracy meant a model was more likely to be fair. Those who opted for higher accuracy noted that they felt higher accuracy meant a model would handle bias better, while those who took the lower accuracy route noted they felt accuracy had to be sacrificed to help guarantee fairness.

Fifteen participants cited making an ``educated guess'' regarding model fairness. Participants backed their educated guesses with the accuracy score, outside resources, background knowledge of machine learning models and how they work, or some combination of the three.
Often accuracy was coupled with some other metric or explanation for model selection but some participants made decisions on the fairness of their models without using any metrics.
For two of 15 educated guesses, the decision was based on the models they evaluated using fairkit-learn or AI Fairness 360 in previous tasks. 
For another two participants, the model that was able to achieve the highest accuracy in the the shortest amount of time was considered to be the most fair.
Participants also made assumptions about the dataset, such as how well distributed it is, to determine model fairness.

\begin{tcolorbox}
Our findings suggest that data scientists sometimes struggle with reasoning
about fairness without the proper tooling. Data scientists have different,
sometimes contradictory, ways of reasoning about fairness when asked to do
so, often using accuracy as a proxy for fairness, despite clear evidence that 
those metrics are often at odds.
\end{tcolorbox}

\section{Threats to Validity}

\paragraph{External validity} Our study compared fairkit-learn to a subset of the tools available for training and evaluating machine learning models. To increase the generalizability of our findings, we selected tools that are considered to be state-of-the-art and supported by industry practices. 

We derived tasks to evaluate our tool, however, the tasks and results may not be representative of what real engineers do and the models they would build. To mitigate the effects of this threat, we provided participants with real datasets that have been used in research and practice. We also asked participants to evaluate bias against attributes that practitioners would care about (e.g., race).

\paragraph{Internal validity} Our user study participants were students completing the study as a homework assignment. This leads to the potential for selection bias. The course included a diverse set of participants and we randomly assigned the task and tool ordering across participants, somewhat mitigating this threat.

The design of our study had participants complete the same tasks for each tool, which introduces the potential for testing bias. To minimize this threat, we had participants use a different dataset for each task which meant different approaches would be needed to meet the goals for each task. 

\paragraph{Construct validity} Given the technical and time requirements for completing the study, one issue we encountered was the effect of technical difficulties and time management of participants on data collection. We included various safeguards for keeping track of participants' contributions and minimizing the effects of technical difficulties on study completion and data collection.

\section{Discussion}

Our findings suggest that tools like fairkit-learn and AI Fairness 360 can
help data scientists find fair models over tools such as scikit-learn (that
are not designed for this purpose). Further, we find that while AI Fairness
360 may be better for focusing on fairness alone, fairkit-learn is able to
help data scientists find the best balance between fairness with other
quality concerns, such as accuracy. This section discusses the implications
of our findings.

\subsection{Using Pareto optimality for balanced models}

One of the primary contributions of fairkit-learn is the ability to search a
large space of models and return results from only the best or optimal models
with respect to the relevant metrics. This is done by calculating the Pareto
optimal set of models from a given set of evaluated models and metrics.
Rather than data scientists having to do their own coding and math to compute a large
number of models to get a sense for where the best balance lies, data
scientists can use the grid search provided by fairkit-learn to find fair models and understand the fairness-quality trade-offs.

While the notion of a Pareto optimal set of models can be useful for finding the most fair models, our study suggests it is most useful for trying to find models that balance more than one metric. In the case of our study,nwe wanted to balance accuracy and fairness. But in the real world, data
scientists may have other metrics they want to balance, including multiple
quality metrics. Fairkit-learn can help data scientists find models that are
Pareto optimal for a given set of metrics, thereby increasing the possibility
of improving the models used for a given dataset or task.

\subsection{The importance of fairness tools}

Findings from our study suggest that even without fairness machine learning
tools, data scientists will sometimes try to reason about the fairness of
machine learning models, often making improper assumptions, leading to poor
reasoning. On the one hand, the first step to training fair models is
thinking about model fairness while training it. On the other hand, one can
argue that ad hoc rationale for model fairness is no better (if not worse
than) not evaluating for fairness at all.

Although some participants found reasonably fair models when using
scikit-learn, more often than not, larger sacrifices had to be made (either
in terms of fairness, or accuracy, or both) when trying to find a
well-balanced model using scikit-learn than when using fairkit-learn and AI
Fairness 360. There was also much more inconsistency behind the rationale for
selecting fair or well-balanced models, which can lead to uncertainty
regarding how fair or unfair a model really is. Our results shed a light on
the importance of using tools that support fair model training and evaluation.

\subsection{Misconceptions regarding model fairness}

When tasked with evaluating model fairness and not equipped with fairness
machine learning tools, participants in our study made various assumptions to
reason about the fairness of a given model. Some of the assumptions
participants made contradicted others, such as the relationship between
accuracy and fairness. Many participants used accuracy as a proxy for
fairness, even though in the datasets available to them, accuracy and
fairness metrics are opposing forces: increasing one typically decreases the
other. Our data suggest that there may be misconceptions data scientists have
regarding what it means for a model to be fair, and, when not armed with
proper tools, make incorrect assumptions and use those assumptions in
evaluating fairness. One reason this discrepancy may exist is due to the
large (and growing) number of ways one can mitigate and measure model
fairness. An important step to training fair models is understanding what it
means for a model to be fair in a given context and what factors may affect
overall fairness or quality of a given model.

\section{Related Work}

We now place fairkit-learn in the context of related contributions on
algorithmic fairness and support for evaluating system fairness and
performance.

\subsection{Perceptions on algorithmic fairness}

While it is important to explore algorithmic fairness when considering the
use of machine learning models, some research has explored how end-users
understand and perceive the notion of algorithmic
fairness~\cite{warshaw2016intuitions, woodruff2018qualitative, grgic2018human}.

Warsaw and colleagues interviewed 21 high school educations individuals on
their beliefs and misconceptions regarding how companies collect data and
make inferences about them using that data~\cite{warshaw2016intuitions}. They
found that most participants believed companies make decisions either based
largely on stereotypes or based on online behaviors and intuition.

Similarly, Woodruff and colleagues conducted an interview study with 44
traditionally marginalized individuals on how they feel about algorithmic
fairness~\cite{woodruff2018qualitative}. When provided with a description of
what it meant for algorithms to be unfair, participants expressed concerns
regarding the implications of algorithmic (un)fairness. They also found that
participants expected companies to address these sorts of things, regardless
of the cause.

Grgi\'{o}-Hla\^{c}a et al.\ propose an explanatory framework to understand
the features people consider fair or unfair to use in decision making and
why~\cite{grgic2018human}. They deployed a series of scenario-based surveys,
developed based on their framework, and found that they can accurately
predict features that would be deemed fair to use.

While these studies help us understand perceptions of algorithmic fairness
and what features may affect the public's perception of a given software's
fairness, it does not help engineers make informed decisions that can ensure,
for example, that the proper features are being considered while also
providing a high performing system. In contrast, fairkit-learn empowers
engineers to explore the space of possible models, with regard to the
features and metrics they care about, such that they can better ensure
algorithmic fairness.

\subsection{Evaluating model fairness \& performance}

Typically, machine learning model performance is evaluated using metrics
pertaining to the accuracy of that model.
scikit-learn~\cite{pedregosa2011scikit} is one of the most common tools used
for training and evaluating machine learning models. scikit-learn is an open
source Python module that provides engineers with a variety of machine
learning algorithms and various metrics for evaluating models for
performance, though no fairness metrics. It is designed to be easy to use and
accessible to non-specialists. While scikit-learn is useful for training and
evaluating models based on their performance, there is no built in
functionality for measuring model fairness or mitigating bias.

There exist tools designed to help engineers reason about fairness in their
machine learning models~\cite{adebayo2016fairml, whatif, Bellamy16}. FairML
helps engineers avoid unintentional discrimination by automatically
determining relative significance of model inputs to that model's
predictions, allowing engineers to more easily audit predictive models.
Meanwhile, Fairway combines pre-processing and in-processing methods to
remove bias from training data and models~\cite{chakraborty2020fairway}.

Google developed the What-If Tool to help programmers and non-programmers
analyze and understand machine learning models without writing
code~\cite{whatif}. Provided a TensorFlow model and a dataset, the What-If
Tool allows you to visualize the dataset, edit individuals in the dataset and
see the effects, perform counterfactual analysis, and evaluate models based
on performance and fairness.

Similar to the What-If Tool is AI Fairness 360, a Python tool suite for
mitigating bias and evaluating models for fairness and
performance~\cite{Bellamy16}. The package includes fairness metrics, metric
explanations, and bias mitigation algorithms for datasets and models. AI
Fairness 360 is designed to be extensible and accessible to data scientists
and practitioners. Also similar is FairVis, a visual analytics system that
supports exploring fairness and performance with respect to certain subgroups
in a dataset~\cite{cabrera2019fairvis}. Similar to fairkit-learn, FairVis
uses visualizations to support this exploration.

While there exists tools that can help engineers evaluate model fairness and
performance, fairkit-learn works with existing tools to help engineers find
Pareto optimal models that balance fairness and performance and a
visualization that makes it quicker and easier to explore the effects of
different model configurations.

\subsection{Training fair models}

\looseness-1
Machine learning approaches that aim to train fair models even when using
biased training data fall into three primary categories: (1)~data
transformation (perturbing input data to quantify bias in data) (2)~algorithm
manipulation (modifying the machine learning cost function typically by
adding fairness constraints or regularization), and (3)~outcome manipulation
(balancing the outcome across multiple groups). Dwork et al.\ formulate
fairness as an optimization problem that can be solved by a linear
program~\cite{Dwork12}. They minimize a loss function while achieving a
Lipschitz property for a defined similarity metric between two individuals
and then they analyze when this local fairness constraint implies statistical
parity. Corbett-Davies et al.\ reformulate algorithmic fairness as
constrained optimization with their fairness definitions as
constraints~\cite{corbett2017algorithmic}. Meanwhile Zhang et al.\ use
adversarial learning as a means for finding fair models~\cite{Zhang18}. Zafar
et al.\ define a measure of decision boundary fairness: the covariance
between sensitive (protected) attributes and the signed distance between the
subjects' feature vectors and the decision boundary of a
classifier~\cite{zafar2017fairness}. They take two different constrained
optimization approaches: (1)~maximizing accuracy subject to fairness
constraints and (2)~maximizing fairness subject to accuracy constraints.
Kamishima et al.\ express fairness regularization as a function of the data
and logistic regression model weights and then they optimize the set of
weights using standard conjugate gradient
methods~\cite{Kamishima2012FairAware}. Their proposed fairness regularization
is differentiable and smooth, thus enabling gradient descent or second order
optimization methods. Thomas et al.\ introduce the Seldonian Framework for
designing machine learning algorithms that perform a one-time safety-check to
produce models that are probabilistically guaranteed to satisfy fairness and
safety constraints, even when applied to unseen data~\cite{Thomas19science}.
Users of algorithms within the Seldonian Framework can apply a large number
of fairness and safety constraints, including multiple simultaneously.
Metevier et al.\ then demonstrate contextual bandit algorithms within the
Seldonian Framework~\cite{Metevier19neurips}, which can satisfy delayed
impact constraints~\cite{Liu18}. While designing novel classification
techniques that explicitly optimize for fairness has shown great promise,
fairkit-learn tackles a related by different problem of helping data
scientists understand the quality-fairness trade-offs and make decisions
about which fairness definitions to use and which models to select in their
specific domains.

Model cards can accompany trained machine learning models to inform users of
benchmark evaluations in certain conditions, which can both disclose the
intended use context and warn users of possible misuses of
models~\cite{Mitchell18}. By contrast, fairkit-learn can produce benchmark
results when model cards are not available, and help uses see fairness
metrics and other parameters relevant to their application domain, which the
algorithm's designers may not have considered a priori.

\subsection{Testing for fairness}

In contrast to correcting for fairness explicitly, there exist a number of
open-source software systems that \emph{test} for fairness. Galhotra et al.\
present Themis, a system that automatically generates test suites to measure
a (1)~a causal definition of fairness (if two individuals differ in only a
single protected attribute then the system recommendation is the same) and
(2)~group fairness~\cite{Galhotra17fse}. FairTest discovers bias bugs, tests
systems for discrimination, and conducts error profiling of machine learning
algorithms~\cite{tramer2017fairtest}, but does not help remove bias. FairML,
an iterative orthogonal transformation process, aims to remove the effect of
a given attribute from a dataset~\cite{Adebayo16}, which creates variants of
datasets, which then need to be explored by tools such as fairkit-learn,
which, in turn, helps explore the entire space of model configurations and
find the ones that satisfy fairness conditions.

\section{Contributions}

We have presented fairkit-learn, a novel open-source toolkit designed to
support data scientists in training fair machine learning models. A
controlled user study showed that students using fairkit-learn produced models that provided a better balance between fairness and accuracy than students using state-of-the-art tools scikit-learn and IBM
AI Fairness 360. Exploring how data scientists approach evaluating fairness
when fairness tools are not available, we found that they struggle, and often
default to using quality metrics, such as accuracy, as a proxy for fairness
(despite the fact that these metrics are often at odds with fairness).
Overall, fairkit-learn is an effective tool for helping data scientists
understand the fairness-quality landscape, and our user study shows promising
results, suggesting that further work improving and evaluating
fairkit-learn with industrial data scientists is worthwhile.

\section*{Acknowledgments}

This work is supported by the U.S. National Science Foundation under grants
CCF-1453474, CCF-1564162, and CCF-1763423.

\balance

\bibliography{fairness,softeng}
\bibliographystyle{plain}

\end{document}